\documentclass[pdflatex,sn-nature]{sn-jnl}

\usepackage{graphicx}%
\usepackage{multirow}%
\usepackage{amsmath,amssymb,amsfonts}%
\usepackage{amsthm}%
\usepackage{mathrsfs}%
\usepackage[title]{appendix}%
\usepackage{xcolor}%
\usepackage{textcomp}%
\usepackage{manyfoot}%
\usepackage{booktabs}%
\usepackage{algorithm}%
\usepackage{algorithmicx}%
\usepackage{algpseudocode}%
\usepackage{listings}%
\usepackage{geometry}
\usepackage{rotating}

\usepackage{arydshln}
\usepackage{lineno}

\begin{document}

\title[Article Title]{Continuous Sleep Depth Index Annotation with Deep Learning Yields Novel Digital Biomarkers for Sleep Health}

\author[1]{\fnm{Songchi} \sur{Zhou}}

\author[2]{\fnm{Ge} \sur{Song}}

\author[3]{\fnm{Haoqi} \sur{Sun}}

\author[4]{\fnm{Deyun} \sur{Zhang}}

\author[5]{\fnm{Yue} \sur{Leng}}

\author[3]{\fnm{M. Brandon} \sur{Westover}}

\author*[1]{\fnm{Shenda} \sur{Hong}}\email{hongshenda@pku.edu.cn}




\affil[1]{\orgdiv{National Institute of Health Data Science}, \orgname{Peking University}}

\affil[2]{\orgdiv{Department of Bioinformatics and Biostatistics}, \orgname{Shanghai Jiao Tong University}}

\affil[3]{\orgdiv{Department of Neurology}, \orgname{Beth Israel Deaconess Medical Center, Harvard Medical School}}

\affil[4]{\orgname{HeartVoice Medical Technology}}

\affil[5]{\orgdiv{Department of Psychiatry and Behavioral Sciences}, \orgname{University of California, San Francisco}}






\abstract{Traditional sleep staging categorizes sleep and wakefulness into five coarse-grained classes, overlooking subtle variations within each stage. It provides limited information about the duration of arousal and may hinder research on sleep fragmentation and relevant sleep disorders. To address this issue, we propose a deep learning method for automatic and scalable annotation of continuous sleep depth index (SDI) using existing discrete sleep staging labels. Our approach was validated using polysomnography from over 10,000 recordings across four large-scale cohorts. The results showcased a strong correlation between the decrease in sleep depth index and the increase in duration of arousal. Specific case studies indicated that the sleep depth index captured more nuanced sleep structures than conventional sleep staging. Gaussian mixture models based on the digital biomarkers extracted from the sleep depth index identified two subtypes of sleep, where participants in the disturbed sleep group had a higher prevalence of sleep apnea, insomnia, poor subjective sleep quality, hypertension, and cardiovascular disease. The disturbed subtype was associated with a 42\% (hazard ratio 1.42, 95\% CI 1.24-1.62) increased risk of mortality and a 29\% (hazard ratio 1.29, 95\% CI 1.00-1.67) increased risk of fatal cardiovascular disease. Our study underscores the utility of the proposed method for continuous sleep depth annotation, which could reveal more detailed information about the sleep structure and yield novel digital biomarkers for routine clinical use in sleep medicine.}





\maketitle

\section{Introduction}

Sleep is essential to human health, and poor sleep poses threats to people's daily life \cite{tregear2009obstructive, smolensky2011sleep} and is linked to numerous diseases \cite{iranzo2015sleep, tsuno2005sleep}. Sleep disorders, including sleep fragmentation and obstructive sleep apnea (OSA), are widespread and associated with adverse health outcomes \cite{martin1996effect,strollo1996obstructive}. In sleep medicine, sleep staging based on polysomnography (PSG) has been an indispensable part of revealing sleep structures for disease diagnosis \cite{boeve2013idiopathic, stephansen2018neural, bohnen2019sleep}. According to the American Academy of Sleep Medicine (AASM) guidelines \cite{berry2012aasm}, sleep and wakefulness are classified into five stages: wake (W), rapid eye movement (REM) (R), non-REM stage 1 (N1), non-REM stage 2 (N2), and non-REM stage 3 (N3) for every non-overlapping 30-second. Despite its importance, manual PSG scoring is labor-intensive and prone to variability \cite{magalang2013agreement, younes2016staging}. Consequently, numerous machine-learning methods have been developed to automate sleep staging with performance comparable to human experts \cite{biswal2017sleepnet, biswal2018expert, perslev2019u, sridhar2020deep, tveit2023automated}. However, current sleep staging is too coarse to accurately reflect sleep depth due to detailed differences in sleep structures within the same stages \cite{uchida1991sigma, uchida1992beta, bonnet1997heart}.

A continuous measure of sleep depth, reflecting the likelihood of being aroused from sleep, holds more value than discrete sleep staging results with respect to research on populations with sleep fragmentation. It has traditionally been studied using external stimuli \cite{johnson1978auditory, gleeson1990influence}. These methods may fail to reflect instantaneous arousal because of the application of stimulus over a long period when sleep depth can already change. Electroencephalogram (EEG) delta power is commonly considered to be closely related to sleep depth. Notably, the odds ratio product (ORP) is introduced as a continuous estimate of sleep depth,  derived from the relationship between EEG power in different frequencies \cite{younes2015odds}. Subsequent studies have validated its effectiveness as an index of sleep depth \cite{meza2016enhancements, qanash2017assessment, younes2017performance, younes2021characteristics, younes2022sleep}. Recent research demonstrates that principal component analysis (PCA) on pre-computed sleep-stage clusters can distinguish sleep stages in a low-dimensional sub-space \cite{metzner2023extracting}. These methods, however, rely solely on EEG data, while AASM recommends using PSG for a comprehensive analysis \cite{berry2012aasm}. Additionally, these measures often involve manually computed features that may not capture nuanced sleep structures. 

Artificial intelligence (AI), particularly deep learning \cite{lecun2015deep}, has become increasingly popular in sleep medicine \cite{goldstein2020artificial, bandyopadhyay2023clinical}. Specifically for sleep depth annotation, AI has the potential to process vast amounts of PSG data and uncover detailed information overlooked by clinicians using some off-the-shelf sleep labels (e.g., sleep staging labels and respiratory events). Given the ordinal nature of Non-Rapid Eye Movement (NREM) stages, expert-labeled staging results can help derive a measure of sleep depth. However, the existing sleep staging labels are coarse-grained as five discrete classes, prohibiting direct supervised training to generate a continuous sleep depth measure. Inspired by the idea of learning to rank widely used in various fields \cite{liu2009learning, karatzoglou2013learning, agarwal2012learning}, we may use ranking-based methods to guide the model to assign higher values to sleep epochs labeled with sleep stages conventionally regarded as deeper sleep. The stages from N1 to N3 represent progressively deeper sleep \cite{patel2022physiology}, fitting well within the learning-to-rank framework. The REM stage, due to its unique significance in clinical settings \cite{ferini2000rem, postuma2010severity, mccarter2012rem}, requires careful handling in comparisons with NREM stages. Nevertheless, REM is deeper than wakefulness and can be integrated into the ranking process, though its comparison with other NREM stages needs caution. Even though the acquirement of the whole-night sleep depth index can serve as a valuable tool to help clinicians inspect the sleep structure from a more detailed perspective, there is no guideline on how to use this whole-night sleep depth index for routine clinical practice for sleep health. Since the whole-night sleep depth index is a type of time series, we can extract several features such as basic time-domain features or complexity-related features as novel digital biomarkers, and investigate their indications for various health conditions. 

\begin{figure}
\centering
\includegraphics[width=0.95\textwidth]{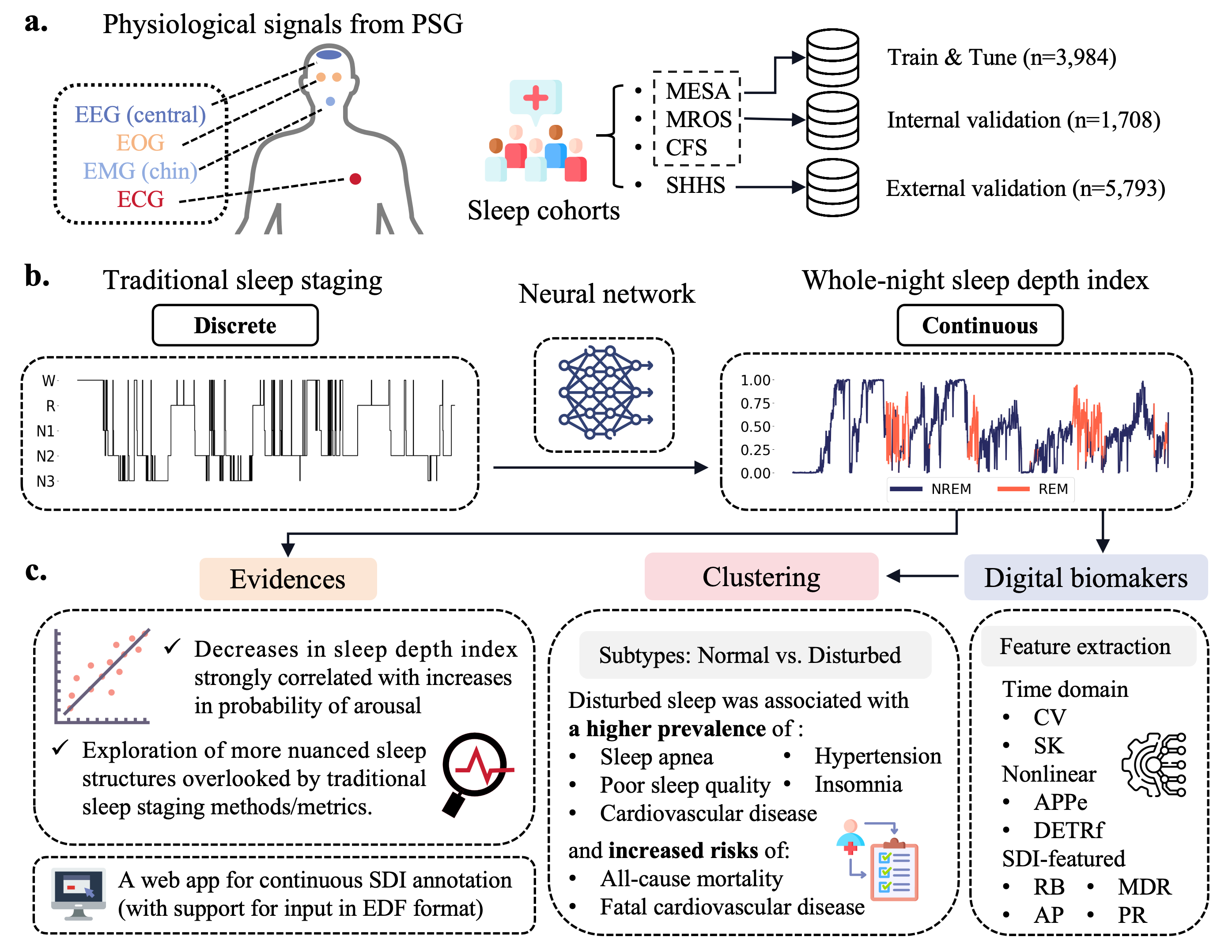}
\caption{General overview of the study. \textbf{a.} Four channels of physiological signals from PSG, EEG, EOG, EMG, and ECG were used in this study. The MESA, MROS, and CFS cohorts were used as the training set and interval validation set and the SHHS cohort was used as the external validation set. \textbf{b.} Using a deep learning method, the neural network was able to transform the discrete sleep staging into a continuous sleep depth index. \textbf{c.} There were several interesting pieces of evidence found in the sleep depth index, which added new insights into the understanding of sleep structure. Digital biomarkers extracted from the sleep depth index were used for clustering, resulting in sleep subtypes exhibiting varied health outcomes. A web app was provided for continuous SDI annotation supporting for EDF-format inputs. PSG, Polysomnography; EEG, Electroencephalography; EOG, Electrooculography; EMG, Electromyography; ECG, Electrocardiography; SDI, Sleep depth index; RB, Ratio below a certain threshold; CV, Coefficient of variation; AP, Proportion of area under the sleep depth index curve; SK, Skewness; MDR, Mean depth value of the REM epoch; PR, Proportion of REM to the total sleep duration; APPe, Approximate entropy; DETRf, Detrended fluctuation analysis.}\label{main}
\end{figure}

To address the aforementioned challenges, we employed a carefully designed pairwise ranking loss to learn the ordinal relations between the NREM sleep stages and between the wake and REM stages, enabling flexible sleep depth annotation within the same stage. This approach will yield a continuous value from 0 to 1, with the larger value indicating deeper sleep for each 30-second PSG epoch. Extensive validations showed that the decrease in sleep depth index is closely associated with the increase in the duration of arousal in the next 30 seconds (Pearson correlation coefficient$>$0.99). Taking the unique role of REM in sleep medicine \cite{ferini2000rem, postuma2010severity, mccarter2012rem} into consideration, we coupled the sleep depth annotation with a REM classification for comprehensive sleep profiling. Then we inspected the nuanced differences in sleep depth index across the same and varied sleep stages, presenting that the same sleep stage could be better distinguished by sleep depth index instead of sleep staging. The whole-night continuous sleep depth index, as a type of time series, allows the extraction of various features as novel sleep biomarkers. We used the Gaussian mixture model to obtain two clusters with the extracted digital biomarkers, namely the normal sleep subtype and the disturbed sleep subtype. Subsequent analyses showcased that the disturbed sleep subtype was associated with several poor health conditions including hypertension, sleep apnea, and so on. Figure \ref{main} displays an overview of the study.

In summary, our contributions are as follows:

\begin{itemize}
    \item We propose a first-of-its-kind deep learning method to annotate the sleep depth index using the PSG data and existing sleep staging labels in an end-to-end way. The model structure supports scalable training on large-scale sleep data. We also deploy an easy-to-use web application for automatic annotation of the sleep depth index.

    \item Experiments on large-scale sleep cohorts and external validations demonstrated the effectiveness of our method. The decrease in the sleep depth index was strongly correlated to the increase in the duration of arousal, and the sleep depth index presented more nuanced sleep structures than conventional sleep staging.

    \item The whole-night sleep depth index time series yielded novel digital sleep biomarkers that were used for clustering. The resultant disturbed sleep subtype was significantly associated with a higher prevalence of several poor health conditions and an increased risk of all-cause mortality and fatal cardiovascular disease.
\end{itemize}

\section{Results}

\subsection{Data curation and deep learning model development}

In this study, we mainly aimed to use a deep learning method to annotate continuous sleep depth using the existing sleep staging labels. Specifically, EEG, Electromyography (EMG), Electrooculography (EOG), and Electrocardiography (ECG) were extracted from the PSG as the input physiological signals for the model. We then converted sleep staging results labeled by the clinicians to ranks to compute the ranking loss and utilized the REM label for the REM classification loss. The MESA, MROS, and CFS cohorts were used as the training set of the deep learning model, comprising 3984 participants, and the other 1708 participants were organized into the internal validation set. The data from the SHHS cohorts was not included in the model training and thus acted as the external validation set. When trained, the deep learning model enables the transition from the discrete sleep stages to the continuous sleep depth index ranging from 0 to 1. We then analyzed the results related to the learned sleep depth index in the following sections.

\subsection{Distribution of sleep depth index and concordance across sleep stages}

Figure \ref{dist} presents boxplots showing the distribution of the sleep depth index across the five sleep stages for each cohort. For the W stage, most sleep depth indices were below 0.1. Conversely, the deepest stage, N3, generally exhibited the highest sleep depth values, consistent with conventional expectations. Interestingly, some high sleep depth values also appeared in the N2 stage and the REM stage. The N1 stage, representing a deeper sleep than wakefulness, showcased sleep depth values mostly ranging between 0.1 and 0.5. Both the N2 and REM stages displayed a wide range of sleep depth indices, indicating varying sleep structures within the same annotated stage. In Table \ref{rem-table}, the first column showed the Spearman's rank correlation coefficient between the sleep depth index and the four sleep stages (W, N1, N2, and N3), which were encoded as progressively deeper sleep transitions (REM stage was not incorporated since its ordinal relation to other NREM stages could not be ascertained). The correlations, all exceeding 0.85, demonstrated good averaged concordance between the sleep depth index and traditional sleep staging results.

To comprehensively showcase whole-night sleep profiles, we have integrated sleep depth annotation with the REM classification. The classification results, detailed in Table \ref{rem-table}, were presented as the area under the receiver operating characteristic (AUROC) values. Overall, the micro-averaged AUROC for the four cohorts was 0.978, with a 95\% confidence interval (CI) ranging from 0.977 to 0.979. For the three internal testing sets, the AUROC values were as follows: 0.990 (95\% CI [0.988,0.991]) for the MESA dataset, 0.984 (95\% CI [0.981,0.986]) for the MROS dataset, and 0.985 (95\% CI [0.981,0.989]) for the CFS dataset. On the external validation SHHS dataset, the AUROC value was 0.975 (95\% CI [0.974,0.976]), demonstrating the robustness of the REM classification part and its ability to generalize well across different datasets. Notably, the third column in Table \ref{rem-table} presents the REM classification without training together with the sleep depth annotation task, where the classification performance was universally slightly worse than those in the second column, suggesting that the joint training model structure could enhance the representation learning of PSG data.

\subsection{Case studies for nuanced sleep structures shown in sleep depth index while not in traditional sleep staging}

\begin{figure}
\centering
\includegraphics[width=1.\textwidth]{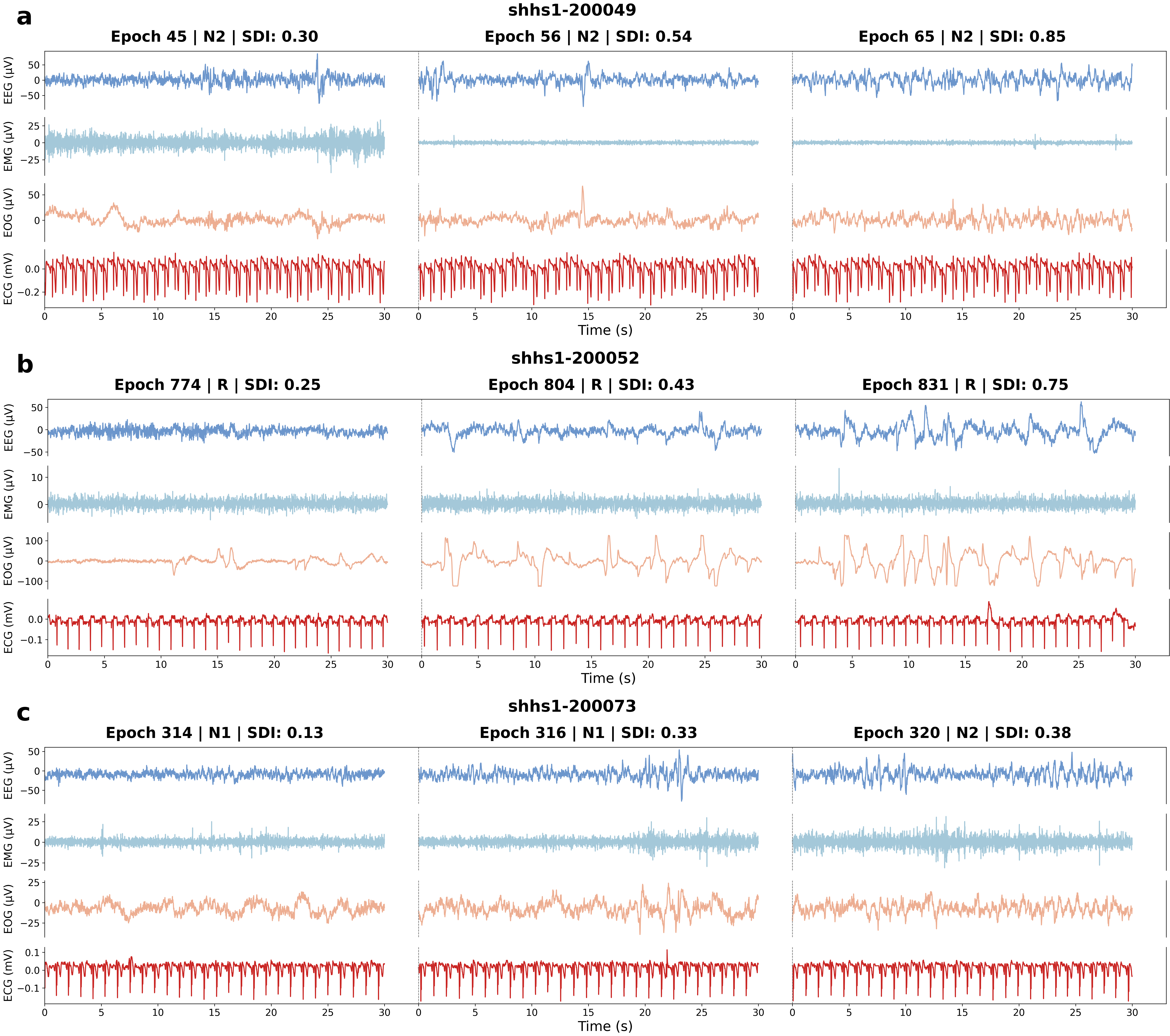}
\caption{Cases of varied patterns of the physiological signals across the sleep stages. \textbf{a.} The three stages belonged to the same N2 stage but the sleep depth values differed. The sleep depth index better captured the lower frequency feature of deep sleep and the smaller magnitude of EMG. \textbf{b.} The three stages belonged to the same REM stage but the sleep depth values differed. The sleep depth index better captured the lower frequency feature of deep sleep and the larger magnitude of EOG. \textbf{c.} The first two stages belonged to the N1 stage but the second one was labeled with a slightly larger sleep depth index. The third stage shared similar patterns with the second stage, but it was labeled as the N2 stage, where the sleep depth values were close. SDI, Sleep depth index; EEG, Electroencephalography; EOG, Electrooculography; EMG, Electromyography; ECG, Electrocardiography.}
\label{case1}
\end{figure}

In this section, we investigated the nuanced sleep structures and patterns captured by the proposed sleep depth index, while overlooked by the traditional sleep staging methods. In Figure \ref{case1}a, the three 30-second epochs were all labeled with the N2 stage by the clinicians. However, apparently different sleep patterns could be noticed. The first epoch featured high-frequency EEG and high-amplitude EMG, resembling a waking state but also showing a representative N2 stage K-complex. The second epoch presented significantly lower EMG amplitude and more low-frequency and high-amplitude EEG waves, indicating deeper sleep states. More low-frequency EEG patterns were observed in the third epoch, which was labeled with a larger sleep depth index. Furthermore, in Figure \ref{case1}b, the three 30-second epochs were all labeled as the REM stage but characterized with varied sleep depth index by our model. We were able to observe more lower-frequency and higher-amplitude EEG features in the epochs with larger sleep depth values. In addition, larger sleep depth indexes were associated with more evident eye movements showcased by the EOG. The N1 stage, which suffers from poor inter-rater reliability, has been challenging for existing sleep staging methods due to the subtle differences between N1 and N2. In Figure \ref{case1}c, although both the first epoch and the second epoch were recognized as N1, the sleep depth index would assign a higher value to the second epoch in relation to the lower-frequency and higher-amplitude EEG pattern. Nevertheless, the physiological signals in the third epoch, which was labeled with the N2 stage, resembled those from the second epoch and thus shared a close sleep depth index.

\subsection{The decrease in sleep depth index and the increase in duration of arousal was highly correlated}

\begin{figure}
\centering
\includegraphics[width=1.\textwidth]{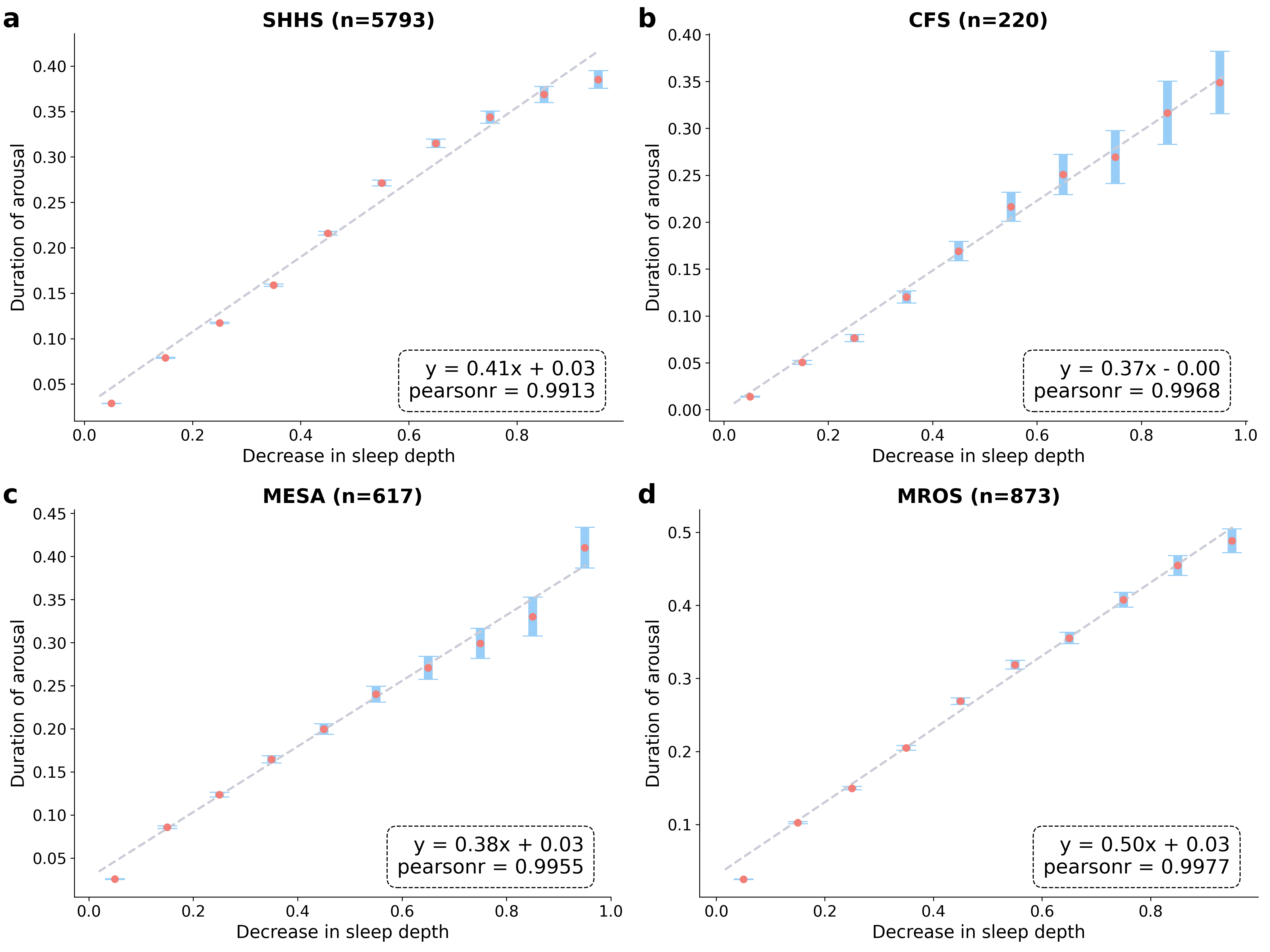}
\caption{The correlation between the decreased magnitude of the sleep depth index and the increase in the duration of arousal. \textbf{a.} The SHHS cohort. \textbf{b.} The CFS cohort. \textbf{c.} The MESA cohort. \textbf{d.} The MROS cohort. The duration of arousal was computed as the proportion of arousal duration in a 30-second epoch. We calculated the ten deciles of 0 to 1 and averaged the values in each interval for linear regression fitting. The red dotted line represented the diagonal line. The average relationship was almost perfectly linear.}
\label{all_scat}
\end{figure}

Arousal during sleep represents a shift from deep sleep to light sleep or from sleep to wakefulness \cite{scammell2017neural}. A low likelihood of arousal indicates deeper sleep depth, making it less likely for the sleeper to be awakened. For the four cohorts studied, arousal events were annotated with their start and duration times. We then defined the duration of arousal for a certain 30-second epoch as the proportion of the arousal duration within that epoch. The decrease in the sleep depth index was computed by subtracting the value at time $t$ from the value at time $t-1$. We created deciles for the annotated sleep depth index, resulting in ten equal intervals. We then averaged the arousal durations within each interval, with error bars representing two-sided confidence intervals. As shown in Figure \ref{all_scat}, the linear regression analysis for each cohort revealed strong correlations between the decrease in sleep depth index and the increase in the duration of arousal. Specifically, the Pearson correlation coefficients were 0.9913 for SHHS, 0.9968 for CFS, 0.9955 for MESA, and 0.9977 for MROS. It could be noted that larger decreases in sleep depth correspond to broader confidence intervals, probably attributed to the fewer PSG epochs with long arousal durations. Moreover, we investigated this linear relation with more bins (100) split shown in Figure \ref{scats2}, where prominent correlations could still be found.

\subsection{Clustering based on the digital biomarkers derived from sleep depth index resulted in two subtypes with different health conditions and outcomes}

As for the whole-night continuous sleep depth index for each subject, we extracted a set of time series features as the novel digital biomarkers. In the time domain, the coefficient of variation (CV) and skewness (SK) were computed. Based on the physiological nature of the sleep depth index, we then devised several intuitive features to reflect the sleep states. The first was the ratio below a certain threshold (RB), indicating the proportion of shallow sleep during the night. We set 0.2 as the threshold in our study. The second one was the proportion of area under the sleep depth index curve (AP), computed by dividing the integration value of the sleep depth index by the total sleep duration. This feature showcased the efficiency of sleep more accurately than the conventional sleep efficiency metric, which was computed as the ratio of sleep duration to the in-bed period. The approximate entropy (APPe) and the detrended fluctuation analysis results (DETRf) were extracted to analyze the complexity dimension of the sleep depth index. Since we have acquired the REM classification results at the same time, the mean depth value of the REM epochs (MDR) and the proportion of REM to the total sleep duration (PR) were also extracted. Subsequently, these digital biomarkers were used as input for the Gaussian mixture model for clustering, resulting in two subtypes of sleep, namely the normal sleep subtype and the disturbed sleep subtype.

\begin{figure}
\centering
\includegraphics[width=1.\textwidth]{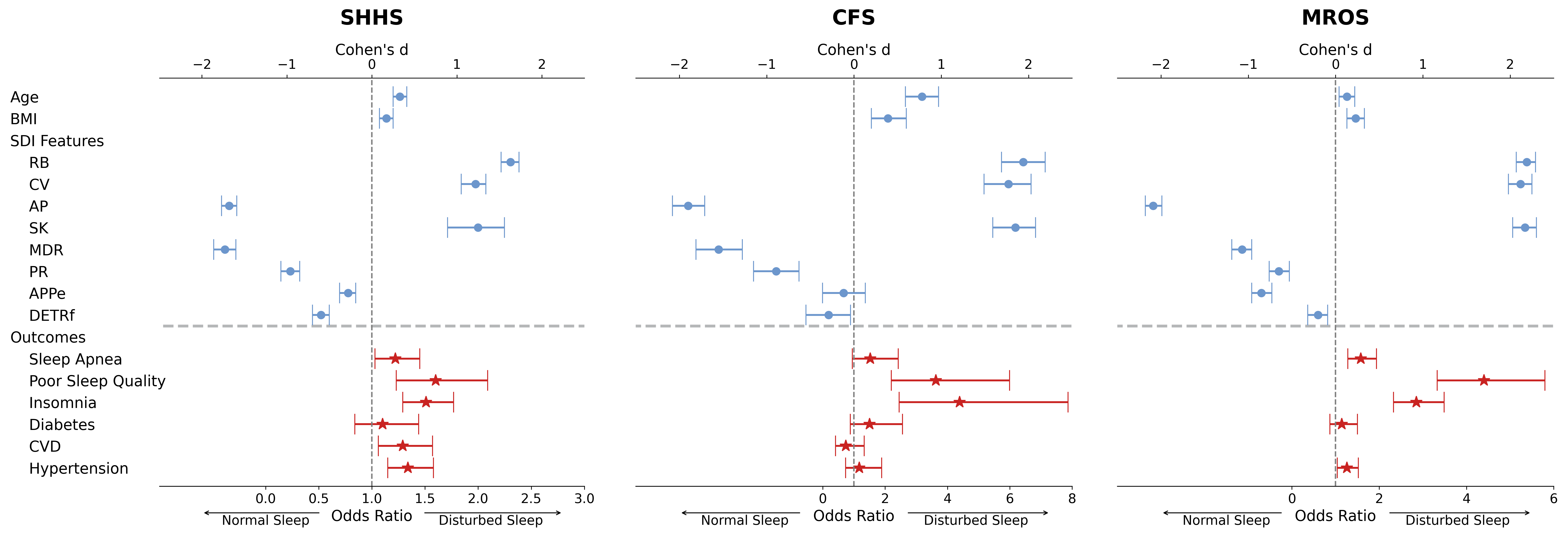}
\caption{Effect size estimates with 95\% CIs for demographic, SDI-based features, and several health outcomes. The region above the dashed line with blue data points was about the comparison of continuous variables, using the t-test to compare the between-group differences and Cohen's d as the measure of effect size. The effect size was computed by misusing the value of the normal sleep group from the disturbed sleep group. The lower region was about the comparison of categorical variables, using the Chi-squared test to compare the between-group differences and odds ratio as the measure of effect size. The disturbed sleep group was regarded as value 1 when computing the effect size. BMI, Body Mass Index; SDI, Sleep depth index; RB, Ratio below a certain threshold; CV, Coefficient of variation; AP, Proportion of area under the sleep depth index curve; SK, Skewness; MDR, Mean depth value of the REM epoch; PR, Proportion of REM to the total sleep duration; APPe, Approximate entropy; DETRf, Detrended fluctuation analysis; CVD, Cardiovascular disease.}
\label{forests}
\end{figure}

In Figure \ref{forests}, the effect sizes estimates with 95\% CIs for demographic, SDI-based features, and several health outcomes are displayed. The detailed values of the effect sizes and the mean values were stored in Table \ref{combined-table}. Generally, for the three cohorts, SHHS, CFS, and MROS, participants in the disturbed sleep group were older and had a large Body Mass Index (BMI), but this finding was only prominent in the CFS cohort. As for the SDI features, the disturbed sleep featured larger RB and smaller AP, indicating a more dominant ratio of shallow sleep and lower sleep efficiency than the normal sleep group. The larger CV in the disturbed sleep group showcased more dispersed patterns than the normal and the larger skewness indicated more frequent or extremely high values. The normal sleep group was characterized by larger MDR and PR, presenting deeper sleep in the REM and longer duration of the REM stage. Notably, the normal sleep group presented larger APPe and DETRf, showcasing a higher level of complexity when compared with the disturbed group. There was an apparent opposite trend for the CV feature and the complexity-based features. It was suspected that the CV reflects more on the variation of regular disturbance across the night and that those with disturbed sleep patterns were frequently affected by these interruptions. On the other side, as measures of complexity, APPe and DETRf presented more nuanced sleep structures similar to the heart rate variability (HRV) for ECG data, where reduced HRV has been shown to be associated with some poor health outcomes \cite{kleiger1987decreased, bigger1992frequency}.

For the two subtypes, the differences in the prevalence of several health were investigated in the lower region of Figure \ref{forests} labeled with red. Logistic regression controlling for age, BMI, sex, and race was used to estimate the odds ratio (OR) with 95\%CI. Sleep apnea was defined as having an Apnea-Hypopnea Index (AHI) larger than 5. The results of poor subjective sleep quality and insomnia were sourced from the morning surveys of the corresponding cohorts. Note that there was no outcome of cardiovascular disease (CVD) for the MROS cohort. We could see from Figure \ref{forests} that the disturbed group across the three cohorts showcased a significantly higher prevalence of sleep apnea, with the odds ratio ranging from 1.22 to 1.58. Poor subjective sleep quality was also significantly associated with the disturbed subtype, with the odds ratio being 1.60 (95\% CI 1.23-2.09) for the SHHS cohort, 3.63(95\% CI 2.2-5.99) for the CFS cohort, and 4.4 (95\% CI 3.34-5.8) for the MROS cohort. Insomnia occurred more frequently for participants with the disturbed subtypes, where the odds ratios ranged from 1.51 to 4.39. The difference in the prevalence of diabetes between the two subtypes was not statistically significant across the three cohorts. CVD was more prevalent in the disturbed subtype than the normal subtype for the SHHS cohort (OR 1.29 95\% 1.06-1.57), but not statistically significant for the CFS cohort. Populations belonging to the disturbed subtype were associated with a higher prevalence of hypertension for the SHHS (OR 1.34 95\% 1.15-1.58) and MROS (OR 1.26 95\% 1.04-1.52) cohorts.

\begin{figure}
\centering
\includegraphics[width=1.\textwidth]{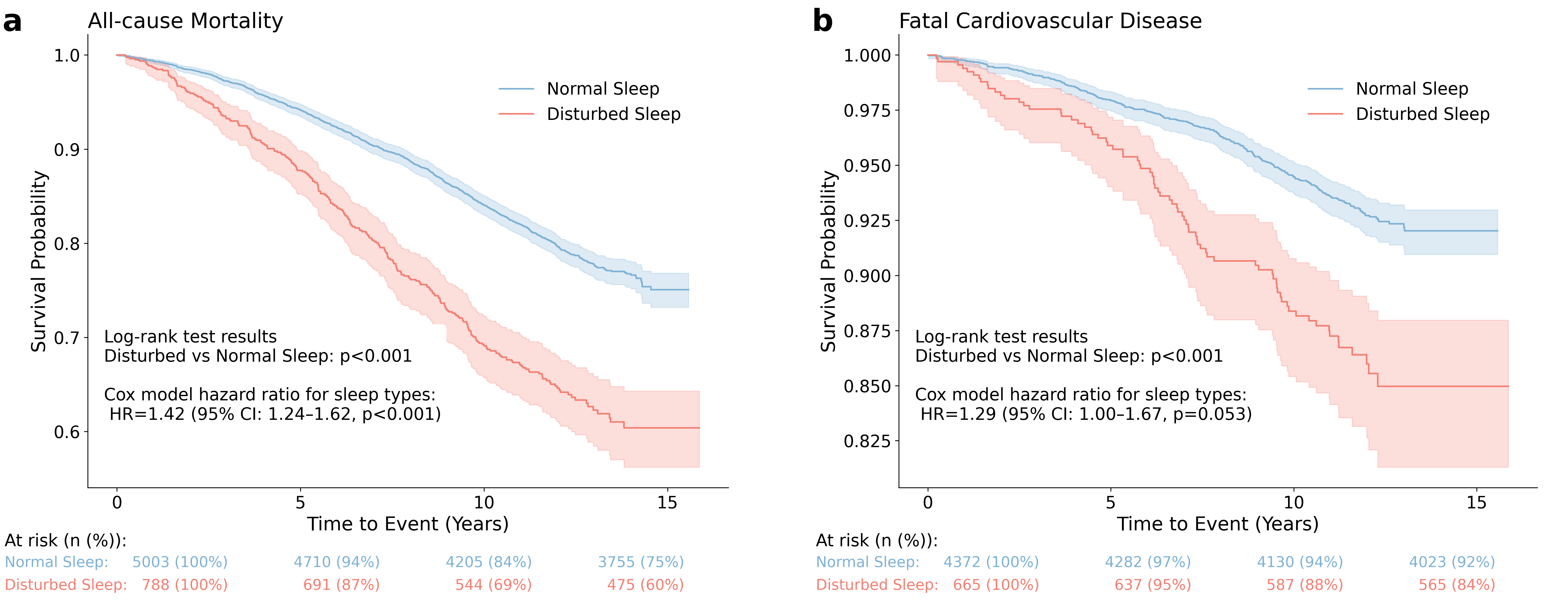}
\caption{Kaplan–Meier curves across the two subtypes for \textbf{a.} all-cause mortality \textbf{b.} fatal cardiovascular disease. HR, hazard ratio; CI, confidence interval.}
\label{kmf}
\end{figure}

As for the SHHS cohort, time-to-event data was extracted for survival analysis of the clustered two subtypes. In Figure \ref{kmf}, Kaplan–Meier survival estimates provided a visual interpretation of the crude probability of all-cause mortality and fatal cardiovascular disease. Log-rank tests showcased significant differences between the two subtypes in the survival probability ($p<0.001$ for both all-cause mortality and fatal cardiovascular disease). From the results, participants in the disturbed subtype had a significantly lower survival probability than the normal subtype. The Cox regression model was then used to determine the hazard ratios (HRs) and 95\% confidence intervals to compare the risks in the normal sleep and disturbed sleep groups, adjusting for age, BMI, sex, and race. For all-cause mortality, the disturbed sleep group was associated with a 42\% (hazard ratio 1.42, 95\% CI 1.24-1.62, $p<0.001$) increased risk compared with the normal sleep group. For the fatal cardiovascular disease, participants with the disturbed subtype suffered a 29\% (hazard ratio 1.29, 95\% CI 1.00-1.67, $p=0.053$) increased risk with respect to the other normal subtype.

\begin{figure}
\centering
\includegraphics[width=.9\textwidth]{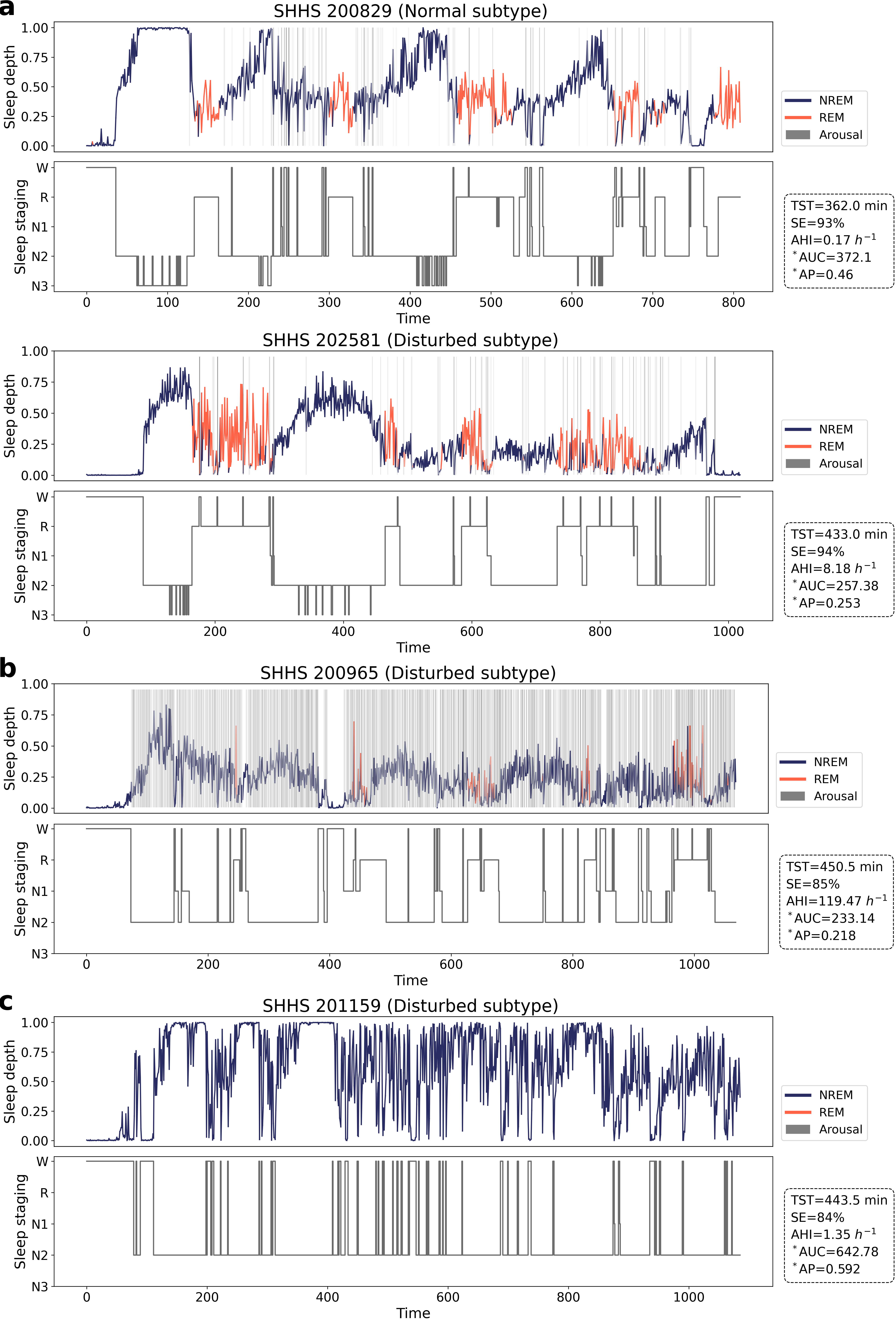}
\caption{Plots of the whole-night SDI along with the expert-labeled hypnogram, where the degree of the gray color indicates the proportion of arousal in a 30-second epoch. In the information box on the right, sleep metrics with * are derived from the sleep depth index. \textbf{a.} Two participants belonging to two different sleep subtypes had close sleep efficiency metrics but significantly different AUC and AP. \textbf{b.} A patient with severe sleep apnea featuring constant shallow sleep and sleep fragmentation. \textbf{c.} The participant with disturbed sleep, though presented a relatively high AP, suffered from excessive sleep depth fluctuation. REM, Rapid eye movement; NREM, Non-rapid eye movement; TST, Total sleep time; SE, Sleep efficiency; AHI, Apnea Hypopnea Index; AUC, Area under the sleep depth index curve; AP, Proportion of area under the sleep depth index curve.}
\label{case2}
\end{figure}


In Figure \ref{case2}, we investigated the differences in SDI patterns for the two clustered subtypes with some instances. We presented the full-night SDI annotations and hypnograms for four SHHS patients. The boxes on the right displayed some metrics such as total sleep time (TST), sleep efficiency (SE), AHI, area under the sleep depth index curve (AUC), and AP. Specifically, for Figure \ref{case2}a, two participants with similar structure hypnograms but belonging to varied clustered sleep subtypes were compared. The participant with disturbed sleep was labeled with a sleep efficiency of 94\%, a little larger than the one displayed in the upper sub-figure of 93\%. However, the latter showcased a larger AP (0.46) than the former (0.253), and in the plots of SDI, we could observe that the participant with disturbed sleep had very shallow sleep in the second half of the night, which would not be discovered by the traditional sleep staging method. In Figure \ref{case2}b, a patient characterized by constant arousal and severe sleep apnea featured shallow sleep across the whole night, ending with a very low AP (0.218) and AUC (233.14) but still a high sleep efficiency (85\%) and high total sleep time (450.5 min). Next for the participant with the disturbed subtype as shown in Figure \ref{case2}c, no REM or N3 were identified, and no sleep cycles were observed. Though the participants had a high AUC and AP, the sleep pattern fiercely fluctuated from deep sleep to shallow and vice versa, which was less obvious in the frequent transitions between the wake and N2 stages as shown in the hypnogram.

\section{Discussion}

In this study, we developed and externally validated a deep-learning method for end-to-end annotation of the sleep depth index using a large scale of 11485 PSG recordings. We first viewed the basic properties of the sleep depth index in the second part of the results, the distribution of SDI and its relations to sleep staging. Then in the next part, we checked the correlation between the SDI and the arousal event, which is an indispensable element when researching the sleep fragmentation problem. In the fourth part, we investigated the specific cases showing that SDI would be better than sleep staging in capturing certain nuanced sleep patterns. Finally, in the last section of results, a range of comparisons were made to validate that digital biomarkers extracted from the whole-night SDI had the potential to promote clustering and that different subtypes were associated with significantly varied health conditions and outcomes.

There were some trends of SDI that showcased intuitive findings with respect to the sleep stages. For instance, SDIs in the W and N1 stages were generally small and those in the N3 stages were significantly large. The difference in SDI for epochs in the N2 stage varied drastically, and sometimes the boundary between N1/N2 and N2/N3 was not apparent but the SDIs were close. Unlike traditional sleep staging, which assigns five coarse-grained classes to the sleep state, the proposed sleep depth index offers a more nuanced description of sleep structure, capturing varied patterns of physiological waveform that are not discernible through conventional sleep staging, as shown in Figure \ref{case1}. This variability of sleep depth index is meaningful as it reflects the intricate transitions in sleep states, as demonstrated in our case studies. Given the distinctiveness of the REM stage \cite{ferini2000rem, postuma2010severity, mccarter2012rem}, we integrated sleep depth annotation with REM stage classification, producing comprehensive sleep profiling. The predicted REM would also serve as a supplement to feature engineering in addition to the SDI. The proposed SDI was not purposed to replace the sleep stages completely but acted as a valuable source for sleep clinicians to use. 

In sleep medicine, arousal indicates a transition from deep to light sleep, making it a suitable candidate for measuring sleep depth. Our experiments showed that a decrease in the sleep depth index strongly correlated with an increase in the duration of arousal, suggesting its potential for monitoring sleep disturbance. More importantly, the arousal event was routinely labeled by experienced clinicians in a cumbersome manner and might suffer from inter-rater variability. As an automatic approach, the proposed method presents a more consistent way than human labeling and adds new insights to measure sleep fragmentation. 

Extracting the whole-night SDI for an individual results in a time series rich with information about sleep conditions. Time-domain and Nonlinear features could be directly computed but we could mine more intuitive ones such as AP, which is capable of introducing more nuanced findings than the classic sleep efficiency metric. Based on these digital biomarkers derived from SDI, we performed clustering to obtain two sleep subtypes featuring significant patterns in sleep health. Participants in the identified disturbed subtype were more associated with bad health conditions such as CVD, hypertension, sleep apnea, and insomnia. The disturbed sleep subtype had a higher prevalence of all-cause mortality and fatal cardiovascular disease, which offered meaningful indications for clinical practice and relevant prevention. Notably, these are not all the digital biomarkers we can extract. We then envision that more subsequent studies could be conducted to investigate the effects of various time-series features that are able to be explored by the sleep depth index, thus yielding more novel digital biomarkers for sleep medicine. These digital biomarkers would not totally make the traditional sleep metrics such as TST and SE displaced but would help to explore more sleep patterns that were previously overlooked.

Although the concept of sleep depth is frequently discussed in both public and clinical contexts, there are no widely accepted quantitative measurements for sleep depth. Traditional sleep medicine primarily considers the N3 stage as deep sleep \cite{wolpert1969manual} but a single discrete class is far from enough to precisely reflect a high degree of sleep depth. Recently, the ORP \cite{younes2015odds} index had demonstrated utility in clinical applications \cite{meza2016enhancements, qanash2017assessment, younes2017performance, younes2021characteristics, younes2022sleep}, but it relied solely on EEG and required manual computations by experts, limiting its efficiency and accessibility. Moreover, the method was not open-sourced mainly due to the unavailability of the reference table to rank the EEG power values, hindering the research community from conducting related research. Consequently, based on the abundant off-the-shelf sleep staging labels, we proposed to use deep learning, with its extraordinary performance in clinical medicine \cite{singhal2023large,de2018clinically,van2021deep,courtiol2019deep,ardila2019end,liu2020deep}, for annotating the sleep depth index on large-scale PSG in an end-to-end way. Nevertheless, the native sleep staging labels are coarse-grained and discrete in five classes, prohibiting direct supervised training. To tackle this, we designed a pairwise ranking loss to effectively learn the ordinal relations between stages and take the uncertainty of the relation between the REM stage and other NREM stages into account, making it the first attempt to produce a continuous sleep depth measure from sleep staging labels. The model was based on the Transformer structure \cite{vaswani2017attention}, which is most famous for its scalability of training large-scale neural networks that might show emerging abilities \cite{kaplan2020scaling, dosovitskiy2020image, yang2024biot}. We envision that our proposed method of annotating SDI would add new insights into clinicians' interpretations of the PSG besides the traditional sleep analyses. It could be an important quantitative measurement of sleep depth as it is automatically output by the machine learning model instead of human labeling which might suffer from inter-rater inconsistency.

An easy-to-use web application for automatic annotation of the sleep depth index was deployed online for the evaluation of the polysomnography data. The demonstration is shown in Figure \ref{app}. Users can upload EDF files with labeled channel names to obtain analysis results and visualizations. We are continuously enhancing this application to offer more features.

A limitation of our method is its reliance on four PSG channels (EEG, EMG, EOG, and ECG). Future studies should investigate whether similar performance can be achieved with fewer channels, making the approach more feasible in remote and underserved areas where full PSG monitoring is not available. Previous research has shown the effectiveness of using fewer physiological signals in wearable devices for sleep staging \cite{boe2019automating, radha2021deep}, suggesting the potential for portable sleep depth annotation systems. Although we have mentioned that a better sleep state can be explored in full PSG monitoring, wearable devices may be a good trade-off in certain medical scenarios. The other issue with the proposed method is that we only include a single channel of EEG, EMG, EOG, and ECG in modeling, thus a question comes out as to whether the performance would be further improved by incorporating more channels or taking the respiratory signals into consideration. Moreover, the model could be tested on a larger dataset with an increasing number of parameters to further verify its scalability. Our current model used a 30-second scale for annotation, based on available sleep staging labels. Higher resolutions could be achieved by modifying model structures and loss functions \cite{perslev2021u, perslev2022automatic}, which is a direction for future research.

In summary, our study underscored the potential of the proposed sleep depth index annotation method as a valuable ancillary tool in clinical sleep medicine. The resultant sleep depth index promises several utilities in real clinic practice and can be explored to yield novel digital biomarkers for sleep health. We hope this work will inspire further research into AI's role in sleep depth annotation and its significance in sleep medicine.

\section{Methods}

\subsection{Dataset and preprocessing}

Four large-scale cohorts were used in this study. From the NSRR website \cite{zhang2018national}, the Sleep Heart Health Study (SHHS) is a multi-center cohort study implemented by the National Heart Lung \& Blood Institute to determine the cardiovascular and other consequences of sleep-disordered breathing \cite{quan1997sleep}. The Cleveland Family Study (CFS) is the largest family-based study of sleep apnea worldwide, which was begun in 1990 with the initial aims of quantifying the familial aggregation of sleep apnea \cite{redline1995familial}. Multi-Ethnic Study of Atherosclerosis (MESA) is an NHLBI-sponsored 6-center collaborative longitudinal investigation of factors associated with the development of subclinical cardiovascular disease and the progression of subclinical to clinical cardiovascular disease \cite{chen2015racial}. MrOS is an ancillary study of the parent Osteoporotic Fractures in Men Study \cite{blackwell2011associations}. For the SHHS and MROS cohorts, we used records belonging to visit-1. The basic statistics are shown in Table \ref{demo-stat}. We used the MNE library \cite{gramfort2013meg} to extract the PSG signals from the raw files and resample them to 100 Hz for concordance. The signals were then trunked to 30-second epochs and saved with the corresponding sleep staging labels and arousal annotation. We chose one EEG (C4) channel, one EMG (chin), one EOR (right eye), and one channel of ECG. We split training sets and internal testing sets for the CFS, MESA, and MROS cohorts by a 7:3 ratio. The SHHS cohort was left as the external validation dataset.

\subsection{Definitions of sleep apnea, sleep quality, insomnia, and mortality events}

For the four datasets, Apnea–hypopnea index (AHI) values, computed as (All apneas + hypopneas with $\geq$ 30\% nasal cannula [or alternative sensor] reduction with $\geq$ 4\% oxygen desaturation) / hour of sleep, were extracted from the data harmonized by the NSRR team. Then sleep apnea was defined as AHI $\geq5$. At the time of these PSG studies, subjects were required to complete some morning surveys with respect to their last night's sleep. Note that there are some surveys inquiring about the sleep habits of the participants but we did not use them since we concentrated on the immediate sleep feeling after the PSG studies. In our experiment, three variables \textit{diffa10, ltdp10, rest10} were selected for the SHHS dataset. Specifically, the \textit{diffa10} variable measured the difficulty of falling asleep, with 0 indicating No and 1 indicating Yes. Thus participants with \textit{diffa10} being 0 were defined as having insomnia disorders. The \textit{ltdp10} variable measured the quality of sleep in terms of light or deep sleep, with five degrees where a value of 1 indicated light sleep and a value of 5 indicated deep sleep. The \textit{rest10} variable measured the quality of sleep in terms of restless or restful, with five degrees where a value of 1 indicated restless sleep and a value of 5 indicated restful sleep. We selected degree 1/2 and degree 4/5 of each index to compare the results. Participants with both the \textit{ltdp10} and \textit{rest10} being less than 2 were defined as having poor sleep quality and those with both the \textit{ltdp10} and \textit{rest10} being larger than 4 were defined as having good sleep quality. As for the CFS dataset, we extracted four variables \textit{easlp, difbak, slpqua, desslp}. The \textit{easlp} variable measured how easy is it for the subject to fall asleep last night, which was rated on a scale of 1-6, with 1 being very easy and 6 being not at all easy. The \textit{difbak} variable indicated the difficulty of falling back to sleep, with value 0 showing No and value 1 showing Yes. CFS participants with \textit{easlp} less equal to 3 and \textit{difbak} being 0 were defined as not having insomnia and those with \textit{easlp} more than 4 and \textit{difbak} being 1 were defined as having the insomnia symptom. The \textit{slpqua} variable measured the subjects' feeling of the overall quality of sleep last night, which was rated on a scale of 1-6, with 1 being extremely refreshing and 6 being not refreshing. The \textit{desslp} variable was about the participants' description of sleep last night, with the value 1 being excellent, 2 being very good, 3 being fair and 4 being poor. Then, participants with \textit{slpqua} being less than 3 and \textit{desslp} less than 2 were defined as having good sleep quality, and those with \textit{slpqua} being more than 4 and \textit{desslp} being more than 3 were defined as having poor sleep quality. As for the MROS dataset, three variables \textit{poxfall, poxqual1, poxqual3} were selected. The \textit{poxfall} variable was a numeric index measuring how long it took for the participant to fall asleep at bedtime last night. MROS participants with \textit{poxfall} more than 40 minutes were defined as having the insomnia symptom. The \textit{poxqual1} variable inquired about the participants' feelings about the sleep being light or deep, with five degrees similar to the \textit{ltdp10} variable of the SHHS dataset. The \textit{poxqual3} measured the feelings about the sleep being restless or restful, with five degrees similar to \textit{rest10} in SHHS. MROS participants with both \textit{poxqual1} and \textit{poxqual3} being less than 2 were defined as having poor sleep quality and those with both \textit{poxqual1} and \textit{poxqual3} being larger than 4 were defined as having good sleep quality. Participant deaths in the SHHS cohort, which were identified and confirmed using multiple concurrent approaches including follow-up interviews, written annual questionnaires, telephone contacts, and so on \cite{punjabi2009sleep}, were sourced from the NSRR dataset. The fatal cardiovascular disease was recorded in parent studies datasets at the NSRR website. The censoring time (days to most recent contact or death) was also extracted accordingly.

\subsection{Definitions of digital sleep biomarkers}

We extracted digital biomarkers from the whole-night sleep depth index from the time domain and the non-linear perspective. As for the time domain, the coefficient of variation and skewness were investigated. Specifically for SDI-featured ones, ratio below 0.2 (RB, the ratio of sleep depth index less than 0.2), and the proportion of area to the total sleep time (AP) were extracted. As for non-linear features, approximate entropy and detrended fluctuation analysis were used. Approximate entropy is a technique used to quantify the amount of regularity and the unpredictability of fluctuations over time-series data. Smaller values indicate that the data is more regular and predictable \cite{richman2000physiological}. The detrended fluctuation analysis is a method for determining the statistical self-affinity of a signal, which has been widely used in the physiological domain \cite{hardstone2012detrended}. 

\subsection{Statistical analysis}

Two-sided t-tests were conducted on the between-group differences of the two clustered subtypes in demographic features and digital biomarkers from the SDI. Cohen's d was used as a measurement of the effect size. Chi-squared tests were utilized to compare the between-group differences in health outcomes. The odds ratio was used to measure the effect size, computed by a Logistic model adjusting for age, sex, BMI, and race. Bootstrapping was implemented to compute the 95\% confidence intervals for the estimates. The log-rank test was used to compare the between-group differences in the survival probability for the two subtypes. The Cox regression model was used to compute the hazard ratios, adjusted for age, sex, BMI, and race.

\subsection{Model structure}

An overview of the model is depicted in Figure \ref{model}. First, the raw input PSG was segmented into a sequence of patches. Specifically, we had an input PSG epoch $x \in \mathbb{R}^{C\times L}$, where $C$ was the number of physiological channels and $L$ was the sequence length. In our study, data from four channels were collected in 30-second epochs at a sampling frequency of 100 Hz for predicting sleep states, thus $C$ was 4 and $L$ was 3000. The sequence of each channel was first split into $N_c$ fixed-size patches with patch size $P$ ($N_c=L/P$). Then the patches from different channels were flattened, yielding a 1D vector composed of $N=C\times N_c$ patches. The flattened patch vector was projected to $D$ dimensions through a trainable linear projection. The outputs of this projection were conventionally referred to as patch embeddings. Following ideas of the original Vision Transformer (ViT) architecture \cite{dosovitskiy2020image}, learnable and randomly initialized positional embeddings were added to the projected patch embeddings to provide the model with information about the position of the patches in the PSG. In addition, we added the channel embeddings since for the PSG every channel has the signal modality varying dramatically and recent works have shown the necessity to take this into consideration \cite{bao2023channel, yang2024biot}. We then prepended a learnable \textit{CLS} token to the patch embeddings to represent the global contextual information learned by the model. The final embedding vectors then served as input of the standard Transformer encoder which consisted of alternating layers of multihead self-attention (MSA) and (multilayer perception) MLP blocks \cite{vaswani2017attention}, where LayerNorm (LN) \cite{ba2016layer} was applied before every block, and residual connections \cite{he2016deep} after every block. 
The standard self-attention mechanism allowed the model to weigh the importance of different patches relative to each other. With $Q$, $K$, $V$ being the query, key, and value matrices linearly projected from the input embedding vector $X\in \mathbb{R}^{N\times D}$, 
\begin{align}
    [Q, K, V] = XU_{qkv},\quad U_{qkv}\in\mathbb{R}^{D\times3D_h}
\end{align}
the attention scores were computed as follows: 
\begin{align}
    A=Attention(Q,K,V)=softmax(\dfrac{QK^T}{\sqrt{D_h}}V)
\end{align}
where $A_{ij}$ was computed based on the respective $Q_i$ and $K_j$ representations and $D_h$ was computed as $D/m$ with $m$ being the number of heads for multihead self-attention. Specifically as for the multihead self-attention, it was an extension of the native self-attention in which $m$ self-attention operations (heads) are conducted. Then these outputs were concatenated and projected to the $D$ dimension
\begin{align}
    MSA(X)=[A_1(X);A_2(X);...;A_k(X)]U_{msa}, \quad U_{msa}\in\mathbb{R}^{m\cdot D_h\times D}
\end{align}

The MLP block consisted of two linear layers with a GELU non-linearity. So the $l$th encoder function could be written as
\begin{align}
X'_l&=MSA(LayerNorm(X_{l-1}))+X_{l-1}, &\quad l=1...L_T \\
Encoder(X_l)&=MLP(LayerNorm(X'_l))+X'_l, &\quad l=1...L_T
\end{align}
where $L_T$ was the number of layers of the Transformer encoder. In this study, patch size was set to 100, projection dimension $D$ to 512, encoder depth to 6, heads number to 8, and MLP dimension to 2048. After passing through the transformer block, the encoded \textit{CLS} embedding was separately fed into a MLP for predicting the sleep depth and another MLP for predicting the REM stage, both of which resembled the Transformer's MLP block. The outputs of the sleep depth MLP were scalars with continuous values annotating the sleep depth, and the outputs from the REM MLP were vectors in length of 2 classifying whether the input epoch represented a REM stage. Note that during training the outputs of the depth head were not bounded and we post-processed them with a Sigmoid function to make them have values ranging between 0 and 1. The reason why we did not bound the range in the depth head was that we observed better results obtained in this experiment setting. The pair rank loss and cross-entropy loss were computed on the sleep depth annotation and REM binary classification respectively with details in the following section.

\subsection{Loss function}

For the design of the loss function for sleep depth annotation, we aimed to leverage the ordinal relationships between different sleep stages, while accounting for uncertain relationships between specific sleep stages. Given a batch of predicted sleep depths $\mathbf{p} = [p_1, p_2, \ldots, p_n] $ and corresponding true labels $\mathbf{y} = [y_1, y_2, \ldots, y_n]$, the goal was to minimize the loss that considered the margin between different pairs of sleep stages and penalizing the for incorrect ordinal relations. Let $\mathcal{M}$ be the mapping from pair types to margins. For a pair of sleep stages $(i, j)$, the margin was denoted by $\mathcal{M}_{ij}$ and computed as: 
$$
\mathcal{M}_{ij} = 
\begin{cases} 
1 & \text{if } (i, j) = (0, 1) \text{ or }(i, j) = (1, 0), \\
0.5 & \text{if } (i, j) = (1, 2) \text{ or }(i, j) = (2, 1), \\
1.5 & \text{if } (i, j) = (2, 3) \text{ or }(i, j) = (3, 2), \\
1.2 & \text{if } (i, j) = (0, 4) \text{ or }(i, j) = (4, 0), \\
\end{cases}
$$
$\mathbf{y} \in \{0, 1, 2, 3, 4\}^n$ was the true sleep stage label for the batch of samples, where 0/1/2/3/4 corresponded to W/N1/N2/N3/R. The set of uncertain relationships was denoted by $\mathcal{U}$. First, we needed to compute all possible pairs from the predicted depths and true labels, where $\mathbf{p}_\text{pairs} = \{(p_i, p_j) \mid i \neq j\}, \quad \mathbf{y}_\text{pairs} = \{(y_i, y_j) \mid i \neq j\}$
with each pair represented as $(p_i, p_j)$ and $(y_i, y_j)$. Second, for each pair of true labels, we retrieved the corresponding margins, $\mathcal{V}_{ij} = \mathcal{M}(y_i, y_j) \text{ for } (y_i, y_j) \in \mathbf{y}_\text{pairs}$. In this study, uncertain relationships were identified as $\mathcal{U} = \{(1, 4), (4, 1), (2, 4), (4, 2), (3, 4), (4, 3)\}$ since we were uncertain about the relation between the REM stage and the other three NREM stages. The mask was formulated as an indicator function as $\mathbb{I}[(y_i, y_j) \in \mathcal{U} \text{ for } (y_i, y_j) \in \mathbf{y}_\text{pairs}]$.

Penalties were computed based on the predicted depths and margins:  
$$
\mathcal{P}=max(0, V_{y_iy_j} - sgn(y_i - y_j)(p_i-p_j))
$$
where sgn was the sign function. We then computed the final pair rank loss by averaging the penalties, ignoring the uncertain relationships:
$$
\mathcal{L}_{rank} = \frac{1}{\mathcal{N}} \sum_{(i, j) \in \mathbf{y}_\text{pairs}} \mathcal{P}_{ij} \cdot (1 - \mathbb{I}[(y_i, y_j) \in \mathcal{U}])
$$
, where $\mathcal{N}=\sum_{(i, j)\in \mathbf{y}_\text{pairs}}(1 - \mathbb{I}[(y_i, y_j) \in \mathcal{U}])$

On the other side, we used the cross-entropy loss for the REM prediction task, which was computed as 
$$
\mathcal{L}_{clas}=-\sum_i y_ilog(\hat{y_i})
$$
where $y_i$ was the true label and $\hat{y_i}$ was the corresponding predicted result from the model. The final loss function for the joint model was formulated as:
$$
\mathcal{L} = \mathcal{L}_{rank} + \alpha\mathcal{L}_{clas}
$$
where $\alpha$ was a hyperparameter of loss weight. The overall training objective was to minimize the combined loss $\mathcal{L}$. In this study, the loss weight $\alpha$ was set to 1. The predicted sleep depth index would then be input to a Sigmoid function to make the range between 0 and 1.

\subsection{Evaluation metrics}

The Pearson correlation coefficient was used to measure the correlation between the decrease in sleep depth and the increase in the duration of arousal. The Spearman's rank correlation was used to measure the correlation between the sleep depth and the W/N1/N2/N3 sleep stages. The area under the receiver operating characteristic (AUROC) was used to assess the classification performance for REM stage classification. Confidence intervals were computed by Bootstrapping.

\section*{Data Availability}
The datasets used in this study are available at https://sleepdata.org/.

\section*{Code availability}
The source code is available at https://github.com/sczzz3/SDI.

\backmatter

\bibliography{sn-bibliography}

\section*{Acknowledgements}

This work was supported by the National Natural Science Foundation of China (62102008, 62172018); Clinical Medicine Plus X - Young Scholars Project of Peking University, the Fundamental Research Funds for the Central Universities (PKU2024LCXQ030); PKU-OPPO Fund (B0202301); CCF-Zhipu Large Model Innovation Fund (CCF-Zhipu202414), the Ministry of Science and Technology of the People's Republic of China (STI2030-Major Projects2021ZD0201900). 

Dr. Westover’s laboratory received support from grants from the NIH (R01NS102190, R01NS102574, R01NS107291, RF1AG064312, RF1NS120947, R01AG073410, R01HL161253, R01NS126282, R01AG073598) and NSF (2014431).

\section*{Competing interests}
The authors declare no competing interests.

\newpage
\begin{appendices}



\section{Extended Data}

\begin{table}[htbp]
\caption{Correlation with sleep staging (SS) results measures the conformability of the sleep depth index with the original staging trends (without the REM stage), in which the values are Spearman's rank correlation coefficients. The second column presents the classification performance w.r.t. the REM stage, which is evaluated by the AUROC(Area Under the Receiver Operating Characteristics). The third column is about the performance of REM classification without being jointly trained with the sleep depth annotation module.}
\label{rem-table}
\centering
\begin{tabular}{@{}cccc@{}}
\toprule
Cohort & Corr. SS (w/o REM) & Clas. REM (AUROC) & Clas. REM w/o SD \\ \midrule
MESA   & 0.864$\pm$0.064 (0.859, 0.869)               & 0.990$\pm$0.016 (0.988, 0.991) & 0.986$\pm$0.018 (0.985, 0.988)  \\
MROS   & 0.863$\pm$0.061 (0.859, 0.867)               & 0.984$\pm$0.037 (0.981, 0.986) & 0.978$\pm$0.041 (0.975, 0.981)  \\
CFS    & 0.881$\pm$0.043 (0.875, 0.887)               & 0.985$\pm$0.030 (0.981, 0.989)  & 0.978$\pm$0.042 (0.972, 0.983)  \\
SHHS   & 0.856$\pm$0.076 (0.854, 0.858)               & 0.975$\pm$0.036 (0.974, 0.976) & 0.967$\pm$0.036 (0.966, 0.968) \\ 
All    & 0.858$\pm$0.073 (0.857, 0.860)                & 0.978$\pm$0.035 (0.977, 0.979) & 0.970$\pm$0.036 (0.969, 0.971) \\ \bottomrule
\end{tabular}
\end{table}

\begin{figure}[h]
\centering
\includegraphics[width=1.\textwidth]{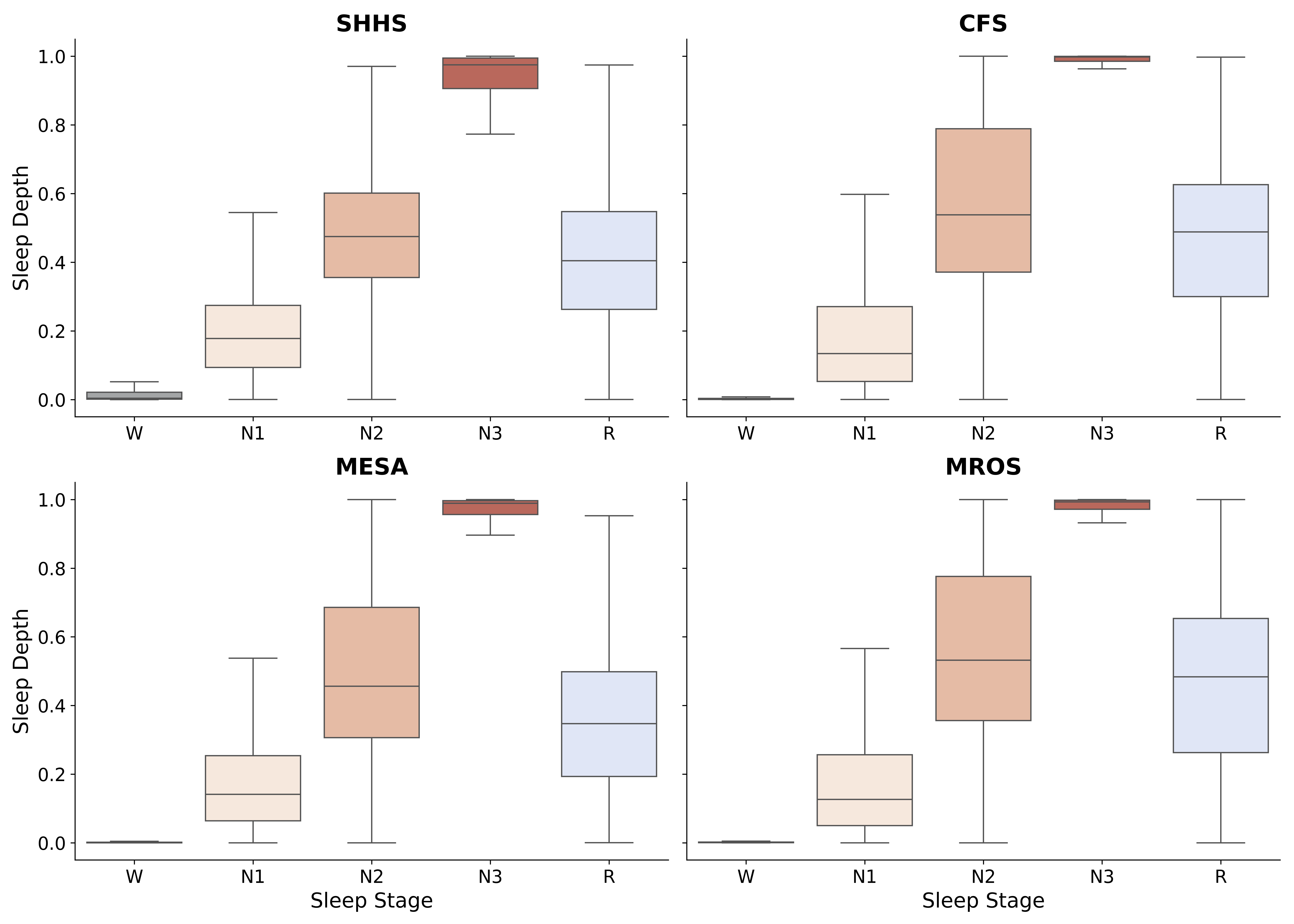}
\caption{Boxplots of the distribution of the sleep depth index across different sleep stages.}\label{dist}
\end{figure}

\begin{figure}[h]
\centering
\includegraphics[width=.9\textwidth]{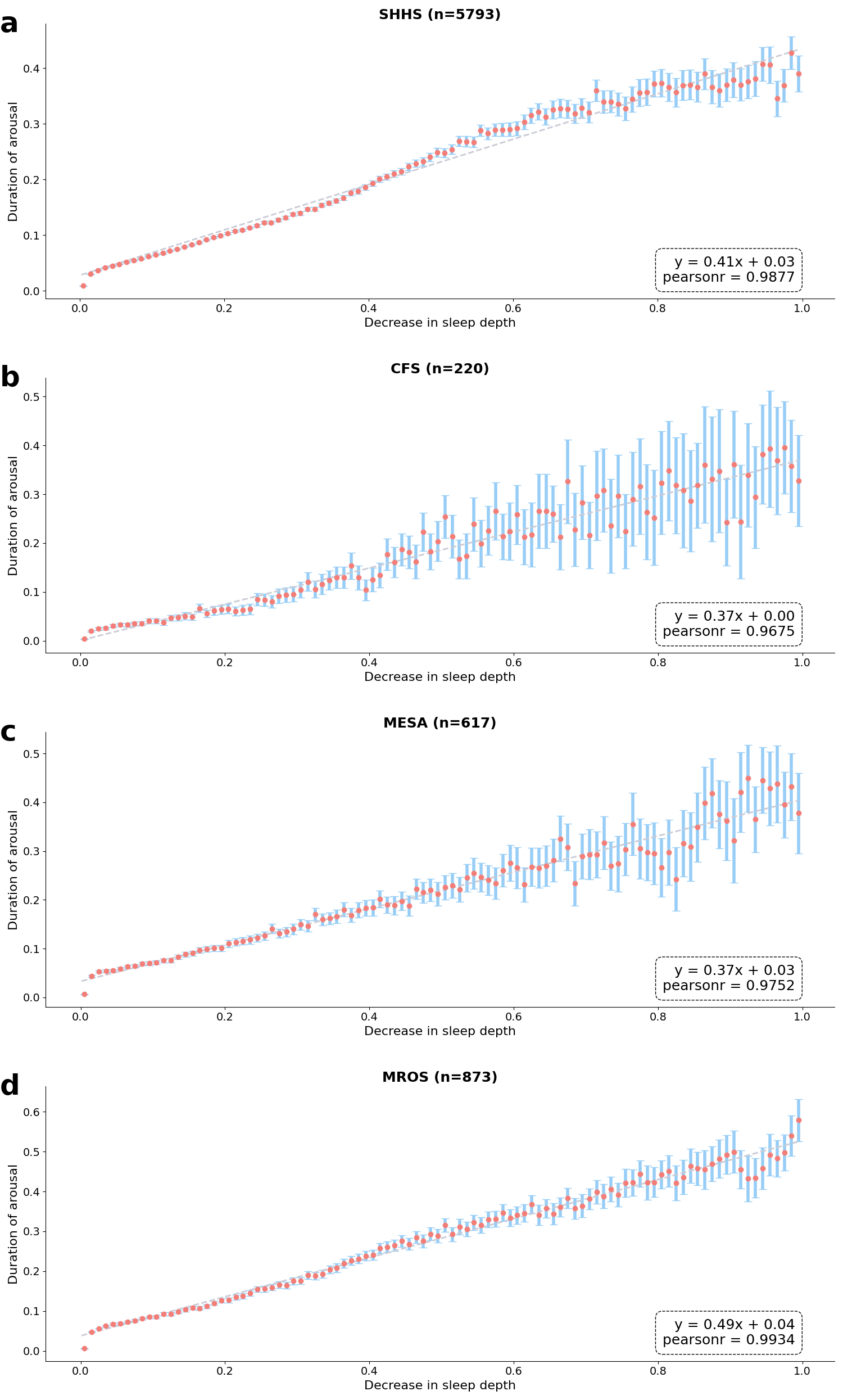}
\caption{The correlation between the decreased magnitude of the sleep depth index and the increase in the duration of arousal with 100 intervals split.}\label{scats2}
\end{figure}

\begin{figure}[h]
\centering
\includegraphics[width=1.\textwidth]{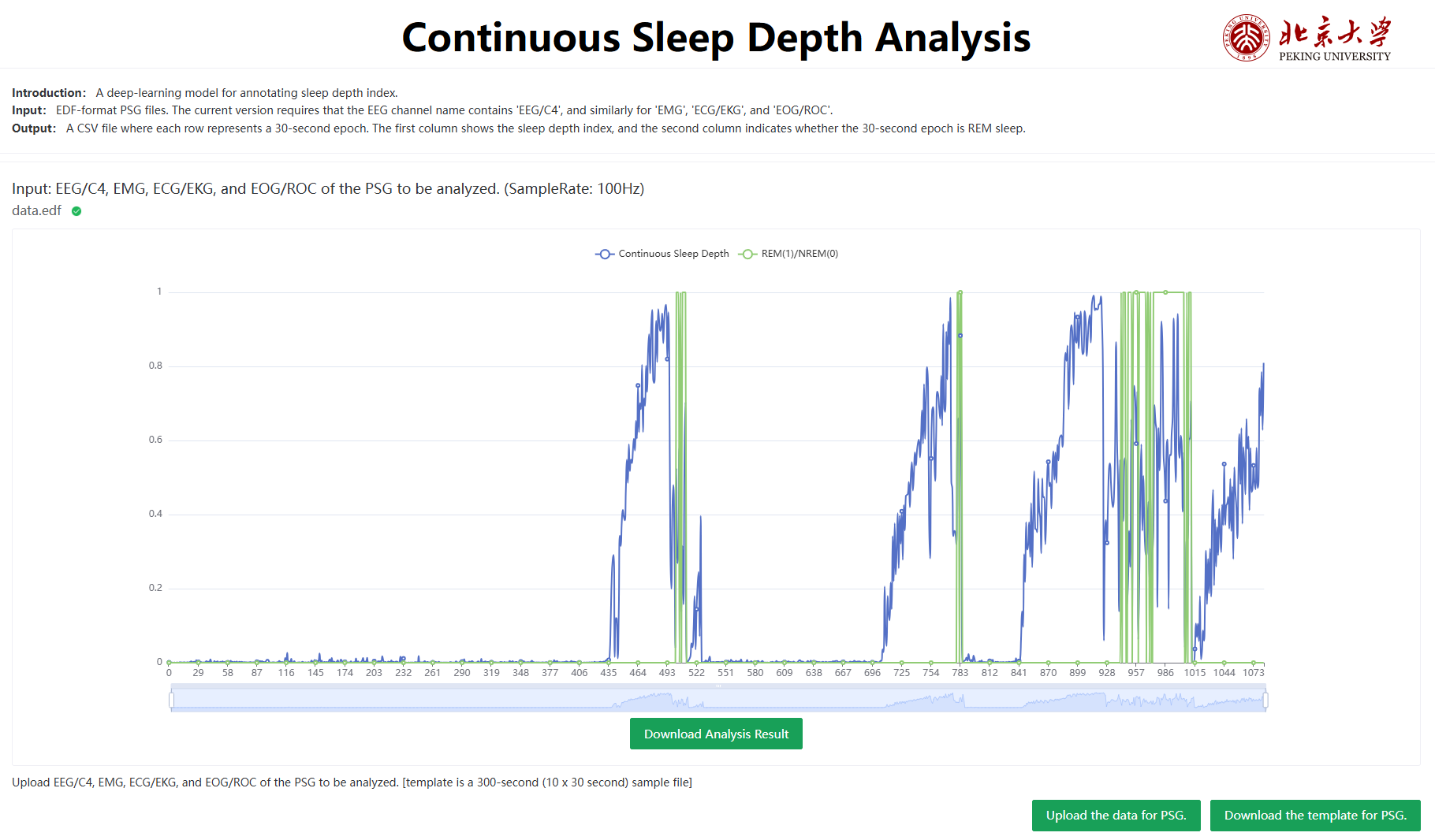}
\caption{The demonstration of the web application for automatic annotation of sleep depth index.}\label{app}
\end{figure}

\begin{table}[h]
\caption{Summary table for the used cohorts}
\label{demo-stat}
\centering
\begin{tabular}{@{}ccccccc@{}}
\toprule
Cohort & Number & Age & Female & BMI & AHI & Timeframe  \\ \midrule
MESA   & 2,055 & 69.4$\pm$9.1 & 54\% & 28.7$\pm$5.5 & 14.8$\pm$16.7 & 2010 - 2013 \\
MROS   & 2,907 & 76.4$\pm$5.5 & 0\% & 27.2$\pm$3.8 & 13.8$\pm$14.4 & 2003 - 2012 \\
CFS    & 730 & 41.4$\pm$19.4 & 55\% & 32.4$\pm$9.5 & 10.2$\pm$18.5 & 2001 - 2006 \\
SHHS   & 5,793 & 63.1$\pm$11.2 & 52\% & 28.2$\pm$5.1 & 10.2$\pm$13.6 & 1995 - 2010 \\ 
\bottomrule
\end{tabular}
\end{table}

\begin{sidewaystable}[htbp]
\setlength{\tabcolsep}{2pt}
\centering
\caption{Demographics, characteristics of SDI features, and health outcomes for two sleep subtypes identified by the clustering. Mean (s.d.) was provided for continuous variables. For continuous variables, t-tests were used to compare the between-group differences, and Cohen's d was used as the effect size. For categorical variables, the Chi-squared tests were used to compare the between-group differences and the odds ratio was used as the effect size. 95\%CI was computed by bootstrapping. N, Normal; D, Disturbed; ES, effect size; SDI, Sleep depth index; RB, Ratio below a certain threshold; CV, Coefficient of variation; AP, Proportion of area under the sleep depth index curve; SK, Skewness; MDR, Mean depth value of the REM epoch; PR, Proportion of REM to the total sleep duration; APPe, Approximate entropy; DETRf, Detrended fluctuation analysis; CVD, Cardiovascular disease.}
\label{combined-table}
\begin{tabular}{@{}lcccclcccclcccc@{}}
\toprule
& \multicolumn{4}{c}{SHHS} & \phantom{abc}& \multicolumn{4}{c}{CFS} & \phantom{abc}& \multicolumn{4}{c}{MROS} \\
\cmidrule{2-5} \cmidrule{7-10} \cmidrule{12-15}
& N & D & ES (95\%CI) & P && N & D & ES (95\%CI) & P && N & D & ES (95\%CI) & P  \\ 
\midrule
\textit{\textbf{Demographics}} \\
Age (years) & 63$\pm$11 & 66$\pm$11 & 0.33(0.25,0.41) & $<$0.001 && 39$\pm$19 & 53$\pm$18 & 0.78(0.59,0.97) & $<$0.001  && 76$\pm$5 & 77$\pm$5 & 0.13(0.04,0.22)& 0.006   \\
BMI (kg$\cdot$m$^{-2}$) & 28.1$\pm$5.0 & 28.9$\pm$5.4 & 0.17(0.09,0.25) & $<$0.001 && 31.7$\pm$9.3 & 35.4$\pm$9.7 & 0.39(0.2,0.6) & $<$0.001  && 27.0$\pm$3.7 & 27.9$\pm$4.3 & 0.23(0.13,0.33)& $<$0.001 \\
\addlinespace
\textit{\textbf{SDI features}} \\
RB & 0.32$\pm$0.11 & 0.51$\pm$0.17 & 1.63(1.52,1.73) & $<$0.001  && 0.4$\pm$0.11 & 0.63$\pm$0.15 & 1.94(1.69,2.19) & $<$0.001  && 0.49$\pm$0.09 & 0.69$\pm$0.1 & 2.19(2.08,2.29)& $<$0.001   \\
CV & 0.78$\pm$0.16 & 1.04$\pm$0.4 & 1.22(1.05,1.34) & $<$0.001  && 0.9$\pm$0.19 & 1.41$\pm$0.56 & 1.77(1.48,2.04) & $<$0.001  && 1.05$\pm$0.17 & 1.54$\pm$0.38 & 2.12(1.98,2.25)& $<$0.001   \\
AP & 0.41$\pm$0.09 & 0.27$\pm$0.09 & -1.68(-1.77,-1.59) & $<$0.001  && 0.43$\pm$0.11 & 0.23$\pm$0.09 & -1.9(-2.08,-1.71) & $<$0.001  && 0.34$\pm$0.08 & 0.19$\pm$0.06 & -2.09(-2.18,-1.99)& $<$0.001   \\
SK & 0.28$\pm$0.33 & 0.91$\pm$1.09 & 1.25(0.89,1.56) & $<$0.001  && 0.21$\pm$0.47 & 1.32$\pm$1.01 & 1.85(1.58,2.08) & $<$0.001  && 0.54$\pm$0.34 & 1.46$\pm$0.64 & 2.17(2.03,2.29)& $<$0.001   \\
MDR & 0.42$\pm$0.08 & 0.25$\pm$0.17 & -1.73(-1.86,-1.6) & $<$0.001  && 0.48$\pm$0.12 & 0.28$\pm$0.19 & -1.55(-1.82,-1.29) & $<$0.001  && 0.48$\pm$0.11 & 0.35$\pm$0.17 & -1.07(-1.19,-0.96)& $<$0.001   \\
PR & 0.2$\pm$0.06 & 0.14$\pm$0.1 & -0.96(-1.07,-0.85) & $<$0.001  && 0.19$\pm$0.06 & 0.13$\pm$0.1 & -0.89(-1.14,-0.63) & $<$0.001  && 0.2$\pm$0.06 & 0.16$\pm$0.09 & -0.65(-0.76,-0.54)& $<$0.001   \\
APPe & 0.99$\pm$0.18 & 0.93$\pm$0.27 & -0.28(-0.38,-0.18) & $<$0.001  && 0.73$\pm$0.19 & 0.71$\pm$0.27 & -0.12(-0.35,0.12) & 0.329  && 0.79$\pm$0.16 & 0.65$\pm$0.21 & -0.85(-0.95,-0.73)& $<$0.001   \\
DETRf & 1.17$\pm$0.07 & 1.13$\pm$0.11 & -0.6(-0.7,-0.5) & $<$0.001  && 1.18$\pm$0.07 & 1.16$\pm$0.11 & -0.29(-0.54,-0.04) & 0.026  && 1.18$\pm$0.07 & 1.17$\pm$0.1 & -0.2(-0.32,-0.09)& 0.001   \\
\addlinespace
\textit{\textbf{Outcomes}} \\
Sleep Apnea &  &  &  &&  &  &  &&  &  &  \\
\hspace{20pt}No & 2527 & 2478 & 1.22(1.03, 1.45) & $<$0.001  && 403 & 48 & 1.52(0.95, 2.42) & $<$0.001  && 839 & 145 & 1.58(1.28,1.95)& $<$0.001   \\
\hspace{20pt}Yes & 296 & 492 & & && 199 & 80 & & && 1482 & 441 & & \\
Poor Sleep Quality &  &  & & &&  &  &  &&  &  & & \\
\hspace{20pt}No & 1006 & 148 & 1.60(1.23, 2.09) & 0.003  && 244 & 31 & 3.63(2.2, 5.99) & $<$0.001  && 767 & 106 & 4.4(3.34,5.8)& $<$0.001   \\
\hspace{20pt}Yes & 571 & 126 & & && 154 & 70 & & && 298 & 183 & & \\
Insomnia &  &  &  &&  &  &  &&  &  &  \\
\hspace{20pt}No & 3536 & 487 & 1.51(1.29, 1.77) & $<$0.001  && 337 & 47 & 4.39(2.45, 7.87) & $<$0.001  && 1941 & 372 & 2.85(2.33,3.49)& $<$0.001  \\
\hspace{20pt}Yes & 1469 & 301 & &  && 57 & 33 & & && 380 & 214 & & \\
Diabetes &  &  &  &&  &  &  &&  &  &  \\
\hspace{20pt}No & 4461 & 655 & 1.1(0.84, 1.44) & 0.001  && 521 & 84 & 1.5(0.88, 2.56) & $<$0.001  && 1760 & 429 & 1.14(0.87,1.5)& 0.066   \\
\hspace{20pt}Yes & 328 & 77 & & && 76 & 41 & & && 280 & 88 & & \\
CVD &  &  &  &&  &  &  &&  &  &  \\
\hspace{20pt}No & 3413 & 429 & 1.29(1.06, 1.57) & $<$0.001  && 535 & 104 & 0.74(0.41, 1.33) & 0.02  &&  &  & & \\
\hspace{20pt}Yes & 959 & 236 & & && 65 & 24 & & &&  &  & & \\
Hypertension &  &  &  &&  &  &  &&  &  &  \\
\hspace{20pt}No & 2961 & 362 & 1.34(1.15, 1.58) & $<$0.001  && 441 & 65 & 1.17(0.73, 1.89) & $<$0.001  && 1200 & 257 & 1.26(1.04,1.52)& 0.001   \\
\hspace{20pt}Yes & 2044 & 426 & & && 154 & 62 & & && 1121 & 328 & & \\
\bottomrule
\end{tabular}
\end{sidewaystable}

\begin{figure}
\centering
\includegraphics[width=1.\textwidth]{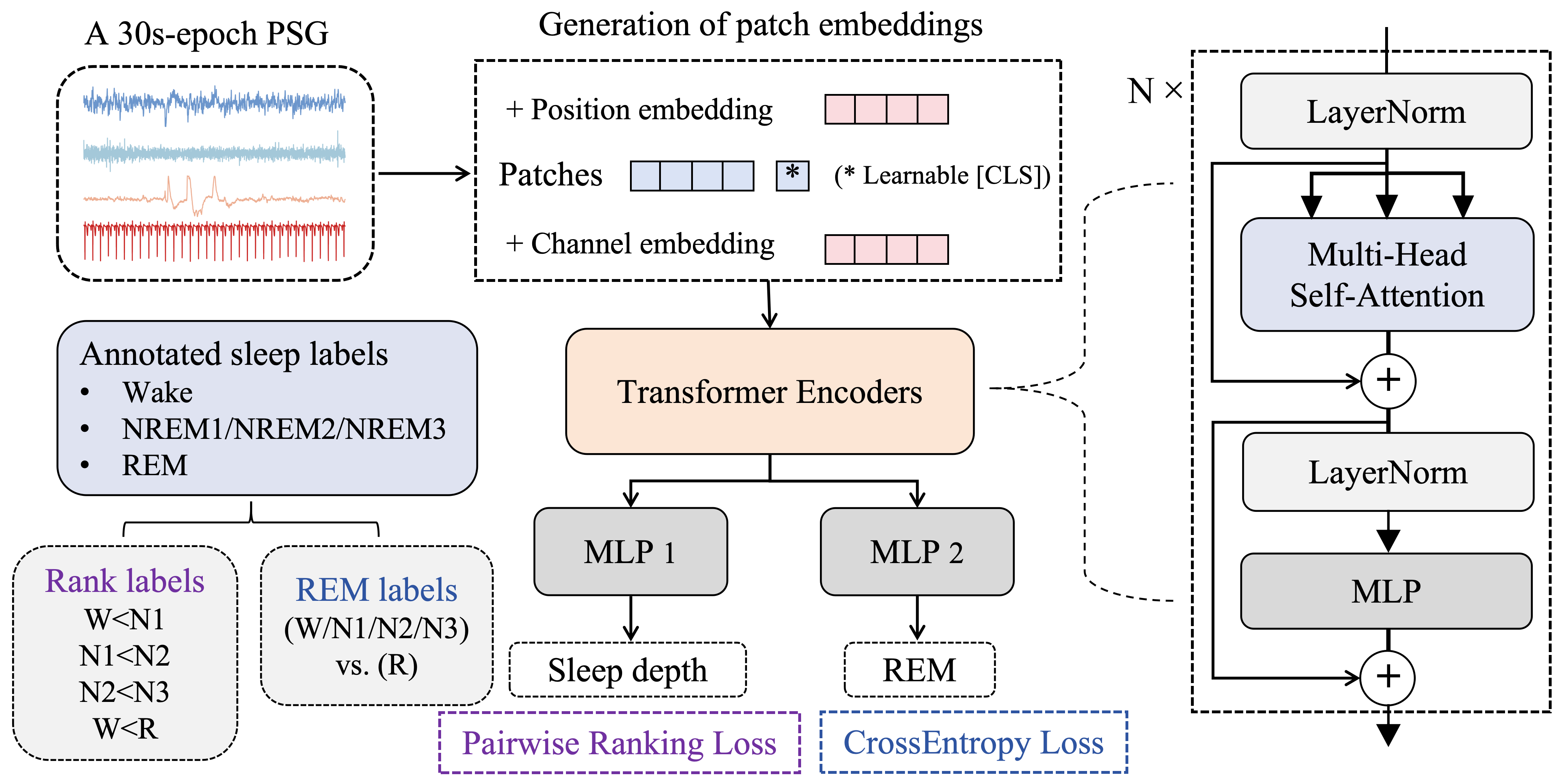}
\caption{An overview of the deep learning model architecture}\label{model}
\end{figure}

\begin{table}[h]
\centering
\begin{tabular}{cc}
\toprule
Abbreviation & Meaning \\ 
\midrule
SDI & Sleep Depth Index \\ 
REM & Rapid Eye Movement \\ 
NREM & Non-Rapid Eye Movement \\ 
PSG & Polysomnography \\ 
EEG & Electroencephalography \\ 
EOG & Electrooculography \\ 
EMG & Electromyography \\ 
ECG & Electrocardiography \\ 
ORP & Odds Ratio Product \\ 
AI & Artificial Intelligence \\ 
ViT & Vision Transformer \\
SHHS & Sleep Heart Health Study \\ 
MESA & Multi-Ethnic Study of Atherosclerosis \\ 
CFS & Cleveland Family Study \\ 
MROS & Osteoporotic Fractures in Men Study \\
AUC & Area Under the Sleep Depth Curve \\
AUROC & Area Under the Receiver Operating Characteristic \\ 
CI & Confidence Interval \\ 
RB & Ratio Below a Certain Threshold \\ 
AP & Proportion of Area Under the Sleep Depth Index Curve \\ 
CV & Coefficient of Variation \\ 
SK & Skewness \\ 
APPe & Approximate Entropy \\ 
DETRf & Detrended fluctuation analysis \\ 
MDR & Mean Depth Value of the REM Epoch \\ 
PR &  Proportion of REM to the Total Sleep Duration \\ 
DETRf & Detrended fluctuation analysis \\ 
AHI & Apnea Hypopnea Index \\ 
CVD & Cardiovascular Disease \\ 
OR & Odds Ratio \\ 
HR & Hazard Ratio \\ 
\bottomrule
\end{tabular}
\caption{Glossary of abbreviations}
\label{abbr}
\end{table}

\clearpage
\section{Analysis of model representations}

\begin{figure*}[hbt!]
\centering
\includegraphics[width=1.\textwidth]{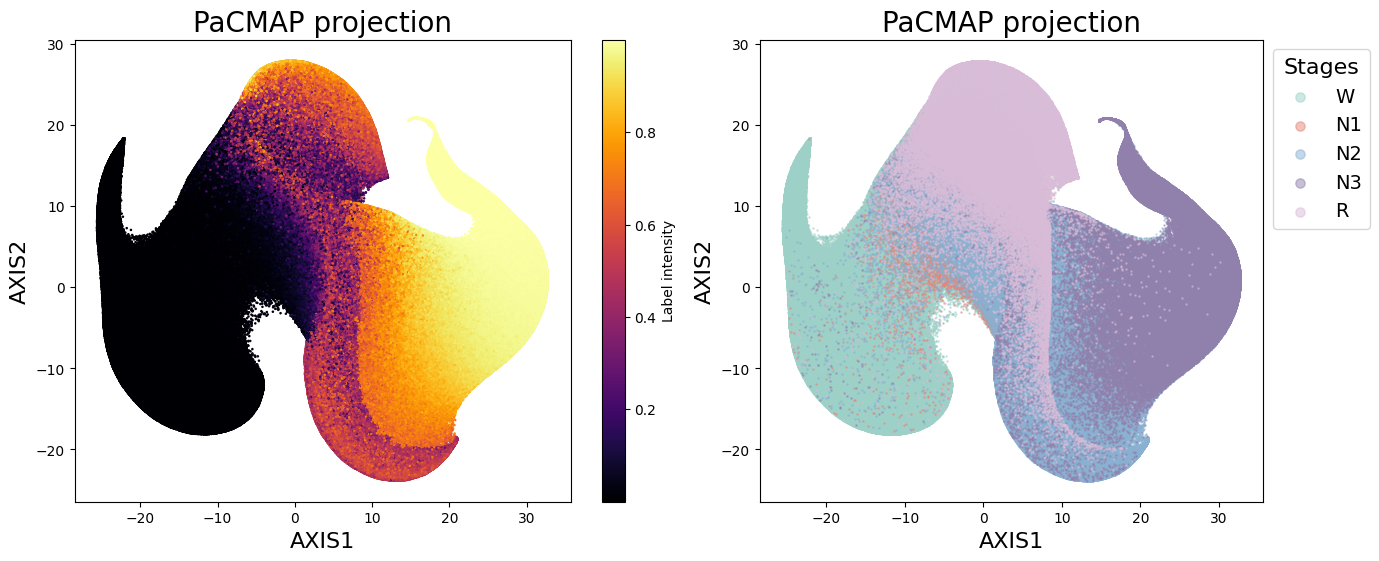}
\caption{PACMAP projection of the features learned by the deep-learning method}\label{pacmap}
\end{figure*}

In this section, we first generated a two-dimensional visualization of the reduced deep-learning features from the last hidden layer representations of the model learned by the PaCMAP algorithm \cite{wang2021understanding}, which was stained with the value of sleep depth, and we also stained the features with the five sleep staging labels. The result is shown in Figure \ref{pacmap}. Specifically, we could find similar patterns shown in Figure \ref{dist}, as the W stage corresponded to a smaller sleep depth index, and the N3 stage corresponded to a larger sleep depth index. The N1 representations were close to W, while the N2 and the REM representations were interleaved in the middle.

\end{appendices}



\end{document}